\documentclass[letterpaper, 10 pt, conference]{ieeeconf}  % Comment this line out if you need a4paper

\IEEEoverridecommandlockouts                              % This command is only needed if
\usepackage{hyperref}       % hyperlinks
\usepackage{url}            % simple URL typesetting
\usepackage{booktabs}       % professional-quality tables
\usepackage{amsfonts}       % blackboard math symbols
\usepackage{nicefrac}       % compact symbols for 1/2, etc.
\usepackage{microtype}      % microtypography
\usepackage{xcolor}         % colors
\usepackage{amsmath}
\usepackage[ruled,linesnumbered]{algorithm2e}
\usepackage[pdftex]{graphicx}
\usepackage{subcaption}

\newcommand{\vc}[1]{\mathbf{#1}}
\newcommand{\vx}{\vc{x}}
\newcommand{\vy}{\vc{y}}
\newcommand{\gG}{\mathcal{G}}
\newcommand{\gV}{\mathcal{V}}
\newcommand{\gE}{\mathcal{E}}
\newcommand{\gN}{\mathcal{N}}
\newcommand{\todo}[1]{}
\newcommand{\junk}[1]{}

\title{\LARGE \bf
Path-Aware Graph Attention for HD Maps in Motion Prediction
}

\author{Fang Da$^{1}$ and Yu Zhang$^{2}$% <-this % stops a space
\thanks{$^{1}$Fang Da is with QCraft Inc,
        {\tt\small fang@qcraft.ai}}%
\thanks{$^{2}$Yu Zhang is with QCraft Inc,
        {\tt\small yu@qcraft.ai}}%
\thanks{Preprint. Accepted to the 2022 IEEE International Conference of Robotics and Automation (ICRA 2022).}
\thanks{Copyright © 2022 IEEE. Personal use of this material is permitted.
Permission from IEEE must be obtained for all other uses, in any current or
future media, including reprinting/republishing this material for advertising or
promotional purposes, creating new collective works, for resale or
redistribution to servers or lists, or reuse of any copyrighted component of
this work in other works.}
}

\begin{document}

\maketitle
\thispagestyle{empty}
\pagestyle{empty}

%%%%%%%%%%%%%%%%%%%%%%%%%%%%%%%%%%%%%%%%%%%%%%%%%%%%%%%%%%%%%%%%%%%%%%%%%%%%%%%%
\begin{abstract}
The success of motion prediction for autonomous driving relies on
integration of information from the HD maps. As maps are naturally
graph-structured, investigation on graph neural networks (GNNs) for encoding HD
maps is burgeoning in recent years. However, unlike many other applications
where GNNs have been straightforwardly deployed, HD maps are
heterogeneous graphs where vertices (lanes) are connected by edges (lane-lane
interaction relationships) of various nature, and most graph-based models are
not designed to understand the variety of edge types which provide crucial cues
for predicting how the agents would travel the lanes. To overcome this challenge,
we propose Path-Aware
Graph Attention, a novel attention architecture that infers the attention
between two vertices by parsing the sequence of edges forming the paths that
connect them. Our analysis illustrates how the proposed attention mechanism can
facilitate learning in a didactic problem where existing graph networks like
GCN struggle. By improving map encoding, the proposed model surpasses
previous state of the art on the Argoverse Motion Forecasting dataset, and won
the first place in the 2021 Argoverse Motion Forecasting Competition.
\end{abstract}

%\text{}

%\begin{keywords}
%Motion Prediction, Graph Attention, HD Maps
%\end{keywords}

%%%%%%%%%%%%%%%%%%%%%%%%%%%%%%%%%%%%%%%%%%%%%%%%%%%%%%%%%%%%%%%%%%%%%%%%%%%%%%%%
\section{Introduction}

In autonomous driving, HD maps are an essential source of information
for the robot, since they capture the semantic structure of the road and thereby
provide driving guidance both as a legal requirement and as a distribution
prior. As the same structural information is understood and respected by human
drivers sharing the road with the robot, the map structure plays a crucial part
in regulating the prediction of motion of other agents. The robot must be
intimately aware of the lanes and traffic controls around it and how each can
affect itself before it can safely navigate them.

The HD maps built for autonomous driving are typically structured as a graph of
map elements, in particular \emph{lanes}, which are connected to one another in
various formations representing admissible ways of traversing the
terrain. For example, two lanes
positioned consecutively in space and connected with a ``sequential'' link
permits driving along them successively, while two lanes side-by-side connected
with a ``lateral'' link enables a ``lane change'' maneuver while driving on one
to reach the other. This highlights that the HD map lane graph is a
\emph{heterogeneous} graph, one with edges of different natures that correspond
to different semantics. Classic graph-based machine learning models, widely
adopted in applications like analysis of social networks
\cite{fan2019graph} and protein-protein interaction \cite{fout2017protein},
often ignore the variety on edge types and focus on processing
features on vertices, assuming the relation between any two vertices is a simple
binary ``connected'' or ``not connected'' relation. This is clearly inadequate
for encoding lane graphs: mistaking a lateral pair of lanes to be
sequential could cause the robot to cut through lanes illegally.

\begin{figure*}
\centering
\includegraphics[width=0.7\linewidth]{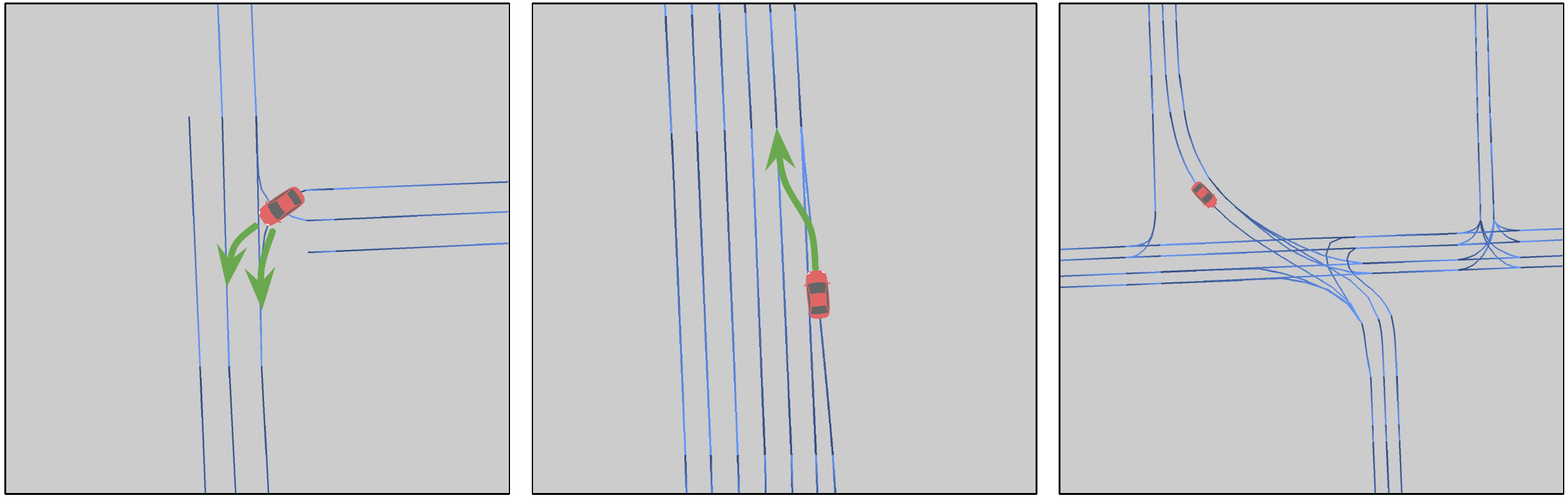}
\caption{Ambiguous road configuration examples from the maps of Argoverse, with
custom rendering to highlight the start (dark-colored) and end (light-colored)
of the lanes. (Left) A turn into a multi-lane road. (Middle) A shallow merge.
(Right) An intersection with several nearby lanes.}
\label{fig:road_configs}
%\myvspace{-3mm}
\end{figure*}

There are techniques available to handle heterogeneity in edges. For example,
one could turn each edge into a vertex with new edges to the original incident
vertices, thus transforming the heterogeneous graph into a larger but
homogeneous bipartite graph. Similarly, one
could directly equip edges with features to capture the same semantic structure.
However, understanding this structure is not easy for the robot: the
complexity of traffic interaction comes not from the range of interaction
types between two adjacent lanes, but from the complexity of
propagating them to nonadjacent but still interacting lanes,
i.e. from their vast array of combinations between the lanes. This is
because in reality there is an enormous variety of combinations of lane
formations in urban areas; the mere vocabulary of a small set of link types
(e.g. sequential and lateral) struggles to
describe more nuanced patterns such as merges/forks and double turns (Fig.
\ref{fig:road_configs}). In order to make map building tractable
and consistent, HD map builders usually have to stick to a small but
well-defined set of link types nonetheless. As a result these more advanced
patterns manifest as combinatoric structures beyond 1-ring neighborhoods, and
the task of dissecting these nonlocal interactions is left to the map encoder
model. Furthermore, the behavior of human drivers often
demonstrate adaptations specific to high order combinations of
these basic patterns, making modeling the full gamut
of lane interactions extremely challenging.

To illustrate the complexity of lane configuration patterns, Fig.
\ref{fig:road_configs} gives a few examples that could potentially be ambiguous
when described by the lane connectivity graph. In the example on the left, a
left turn leads into a multi-lane road. Depending on the speed, road curvature
and traffic, many human drivers would choose to enter the middle or even the
rightmost lane, despite it being illegal in many jurisdictions. This driving
behavior can be explained by hallucinating a lane that does not exist in the map,
or it can be interpreted as entering the neighboring lane of the successor,
thus involving a nonlocal interaction between two lanes with a path of
length 2 in between. The example in the middle contains a long and shallow
merge. During this merge the relation between the two joining lanes transitions
from being neighbors to being nearly coincidental, creating a dilemma for the
labeling of lane neighbor relation. Furthermore, as the red vehicle in the
figure approaches the end of the merge, a left lane change would enter not the
immediate left neighbor lane which is the merge counterpart, but the next lane
over to the left. From the perspective of the lane graph, this would appear to
be an interaction between the red vehicle's lane and its left neighbor's left
neighbor, again a path of length 2. The example on the right is even more
chaotic with several lanes clustered tightly in the middle of the intersection,
and lanes further apart topologically could be interacting.

Based on these observations, we believe an effective encoder model for the HD
map lane graph must have the ability to understand the interaction between
nonadjacent lane pairs from how they come to be linked in between, or in
other words, the ability to infer vertex-vertex attention from the \emph{paths},
or the sequences of edges, between them. Motivated by this insight, the main
contribution of this paper is
a novel attention architecture called \emph{Path-Aware Graph Attention} (PAGA).
As its name suggests, the attention gate between two vertices is learned from
the sequence of edge features, especially the edge types, along the paths
connecting the two vertices, rather than from the vertex features
themselves.

\section{Related Work}

\subsection{Motion Prediction}

Motion prediction has become a central focus in autonomous driving research
over the past few years, and given the importance of HD maps in urban driving,
much research effort has been devoted to effective encoding of the
HD maps. Motion prediction techniques can be broadly categorized according to
the two paradigms of representing spatial features, for encoding maps and
processing interaction with other agents: \emph{rasterized}
and \emph{vectorized}. We briefly review the literature in lieu of these two
categories below with an emphasis on map encoding.

\paragraph{Rasterized} Techniques in the rasterized category interpret the
agent's surroundings as images, typically in the \emph{bird eye view (BEV)}.
Turning the task of semantically understanding the agent's surroundings as a
computer vision problem, these techniques can leverage the vast arsenal of
image-based technologies such as convolutional neural networks. Some authors
argue that a rasterized representation also has the benefit of simplicity in
capturing the agent's spatial context such as the maps \cite{multipath}. A prime
example of this thread of research is the ChauffeurNet \cite{chauffeurnet},
which uses RNN to synthesize predicted trajectories. ChauffeurNet renders the
map, the navigation information and the other objects in a rectangular area in
BEV centered in front of the agent of interest. The roadmap is rendered as an
RGB image containing lane centerlines, curbs \emph{etc}. MultiPath
\cite{multipath} relies on anchor classification and offset regression based on
these context features for producing predictions. A number of works adopt the
same representation for encoding maps, combining it with vectorized
representations for agent motion and interaction: \cite{multimodal_conv} design
a multi-hypothesis FC prediction head on top of the convolutional context
features and the vectorized agent state features, CoverNet \cite{covernet}
dynamically generates anchor trajectories and classify them with similar
features, and Multiple Futures Prediction \cite{mfp} and Multi-Agent Tensor
Fusion \cite{matf} use RNNs to encode and decode agent motion, incorporating the
context features at different stages.

\paragraph{Vectorized} Techniques in the vectorized category, on the other hand,
focus on the topology of the map by treating it as a graph. Compared to raster
images, a graph representation is much more compact and thus enjoys potentially
better efficiency. Depending on the connectivity definition, a graph
representation could also easily describe complicated traffic semantics
such as the case where two lanes are spatially close but prevented from
interacting with each other by a median strip. Notable works in this category
include VectorNet \cite{vectornet} which represents the map as well as the
agent motion as a two-level hierarchy of complete graphs and employs a
self-supervising auxiliary task of completing masked-out vertices,
LaneGCN \cite{lanegcn} which encodes the map as a heterogeneous directed graph
and trains a GCN on it parameterized by edge types, and the follow-up LaneRCNN
\cite{lanercnn} which builds an agent graph to model interaction on top of the
map graph. Our proposed method falls in this category as well.

\subsection{Graph Neural Networks and Attention}

As data in many applications naturally manifest as graphs, deep learning models
for understanding the structure of graphs have long been under active
investigation. \cite{original_gnn} introduce a contraction mapping operator
called Graph Neural Network (GNN) to study the steady state under message
passing between neighbors. This can be viewed as an infinite-horizon recurrent
network; later works \cite{ggnn} make explicit use of RNN for a similar
structure. In order to overcome the difficulty of irregular
neighborhood topology in applying feedforward neural networks to graphs,
PATCHY-SAN \cite{patchy_san} proposes a graph normalization procedure to adapt
neighborhoods of varying sizes to a fixed convolutional layer, and GraphSAGE
\cite{graph_sage} outlines a versatile framework that samples the neighborhood
for a number of vertices, and uses pooling or LSTM to aggregate them.
\cite{spectral_gnn} propose convolutional kernels with the graph spectrum as the
basis, and \cite{gcn} show that under approximation with truncated Chebyshev
polynomials, it is equivalent to the local neighborhood convolution operator,
called Graph Convolutional Network (GCN). MPNN \cite{mpnn} summarizes the various
existing graph-based network structures, including NN4G \cite{nn4g} and GCN, in
a general framework based on the notion of message passing. Graph convolution
has seen wide applications in many domains, ranging from recommendation systems
\cite{ying2018graph} and social networks \cite{kipf2016variational} to
protein-protein interaction \cite{fout2017protein} and
jet physics \cite{henrion2017neural}.

\paragraph{Graph attention}

Popularized in natural language processing \cite{transformer}, the attention
mechanism provides a way to selectively strengthen or weaken the influence of
various elements in the context based on their features, and its natural
adaption to graphs, GAT \cite{gat}, brings an extra layer of expressive power
over graph convolution. Building on top of GAT, GAAN \cite{gaan} adds another
layer of gates to adjust the weight of different attention heads to accommodate
the variability of neighborhood structures. To model long range interactions
beyond the local neighborhood in GAT, SPAGAN \cite{spagan} computes attention
between the vertex of interest and a distant vertex from the features along the
shortest path between the two. This can be seen as an approximation of our
proposed PAGA: as interaction becomes less direct and thus less relevant with
longer path lengths, the shortest path is a zero-th order approximation to the
collection of all paths, from which the path-aware graph attention is computed.

\section{Method}

\subsection{Definitions}

\paragraph{Graphs and models on a graph}
Given a directed graph $\gG = \{\gV, \gE\}$ with vertices $\gV$ and edges $\gE$,
we aim to find network designs that facilitate extracting vertex output features
$\vy: \gV \to \mathbb{R}^{C_y}$ from input features
$\vx: \gV \to \mathbb{R}^{C_x}$
that capture the topological information of the graph as much as possible. We
leave the choice of computation for $\vx$ unspecified here as it is orthogonal
to the discussion on attention. In a graph convolution framework it is common
to use a fully-connected layer from $\vy$ in the previous layer for this
purpose, but other functions can be used too \cite{mpnn}.

This definition works without modification for \emph{heterogeneous graphs}, which are
just graphs whose vertices and edges are heterogeneous, or in other words, are
of different types. The semantic information held in the vertex and edge types
is commonly represented as vertex and edge features; while most graph-based
models directly operate on vertex features, the majority of them are agnostic
of the variability of edge types and edge features, and only consider the binary
connectedness relationship between pairs of vertices, with potentially a
scalar edge weight capturing some notion of interaction intensity. It takes
deliberately designed network architectures to respect the topology of the graph
when processing the semantics of heterogeneous edges.

\paragraph{Attention}
For a vertex of interest $u \in \gV$, an attention mechanism $\Psi(u, v)$ selectively
focuses the ``attention'' on another vertex $v \in \gV$ (that is not necessarily
directly connected to $u$). The attention value modulates how much the features on
$v$ contribute to the results on $u$:
\begin{align}
\label{eq:att}
\vy(u) = \sum_{v \in \gV} \Psi(u, v) \vx(v).
\end{align}

\paragraph{GCN and GAT}
The choice of $\Psi$, in particular what features it depends on, characterizes
the attention mechanism. For example, in GCN \cite{gcn} $\Psi$ is simply the
(renormalized) Laplacian matrix $L = D^{-\frac12}AD^{-\frac12}$
calculated from the degree matrix $D$ and the adjacent matrix $A$ (with a
self-edge added for each vertex), while
in GAT \cite{gat} $\Psi$ is a function of $\vx(u)$ and $\vx(v)$ for $v \in \gN(u)$
where $\gN(u) = \{ v | (u,v) \in \gE \}$ is the 1-ring neighborhood of $u$.
As $\Psi$ typically has a very local support to allow for efficient
implementation, it is common to use multiple layers of such attention structure
to enlarge the receptive field.

\paragraph{Paths}
In the case of heterogeneous graphs, the variable types of edges carry essential
information about the nature of the relations between the vertices in question.
For a vertex pair $(u, v)$, an (edge-) path from $u$ to $v$ of length $l$ is given by
$p = \{e_1, e_2, ..., e_l | e_i \in \gE \}$ satisfying $T(e_{i}) = S(e_{i+1})$ for
$1 \leq i < l$, $S(e_1) = u$ and $T(e_l) = v$, where $S$ and $T$ are the operators
that return the source and target vertex of an edge respectively. There may be
a number of paths of a given length connecting $u$ and $v$, and they collectively
describe the pathways for information to flow between the two vertices through the
graph. In the context of encoding HD maps, such paths may capture how the traffic
in one lane may move into the other, and the sequence of edge types along the
path is a complete descriptor of how the traffic interaction could proceed.

\subsection{Path-Aware Graph Attention}

\begin{figure}[t]
\centering
\includegraphics[width=0.7\linewidth]{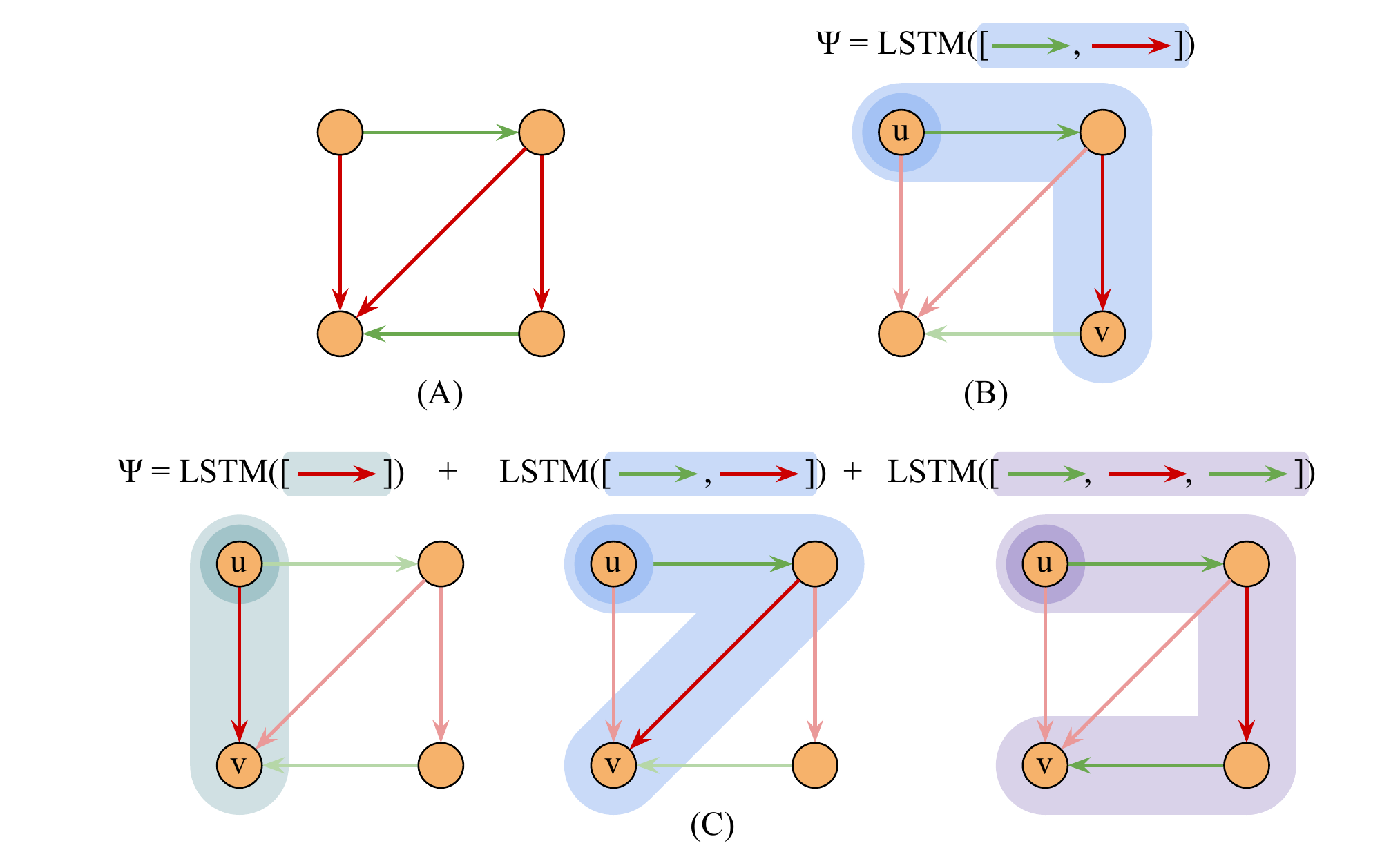}
\caption{Illustration of the computation on edge sequences in PAGA. (A) A
heterogeneous graph with two types of edges (depicted as red and green). (B)
The path connecting $u$ on the top-left corner to $v$ on the bottom-right corner
goes through a green edge followed by a red edge. The edge features of sequence
are encoded by LSTM to produce $\Psi$, the attention between $u$ and $v$.
(C) For a different $v$, there are three paths connecting $u$ and $v$,
all contributing to $\Psi$.}
\label{fig:edge_seq}
\end{figure}

The proposed model, \emph{Path-Aware Graph Attention}, bases the function
$\Psi$ on the edge features $\vx_E: \gE \to \mathbb{R}^{C_e}$ along the
paths connecting $u$ to $v$:
\begin{align}
\Psi(u, v) = \sum_{l \leq \lambda} \phantom{} \sum_{p \in P_l(u, v)} \Phi_l(\{\vx_E(e) | e \in p\}),
\end{align}
where $P_l(u, v)$ is the set of all paths of length $l$ connecting $u$ to $v$,
and $\Phi_l$ is a learnable \emph{feature extractor} function to produce the
attention values from the
sequence of edge features along a path of length $l$, such as a neural network.
$\lambda$ is a hyper-parameter controlling paths up to how long shall be
considered: above a certain length the interaction represented by the
paths is simply too indirect to be relevant, and including them would not improve
the result while still incurring more computation cost (since $|P_l|$ generally
grows with $l$). This attenuation effect with $l$ should arise naturally in the
learned $\Phi_l$ functions, but it can also be explicitly modeled by setting
$\Phi_l(\{\vx_E\}) = \gamma^l \Phi(\{\vx_E\})$ with a $\gamma \in (0, 1)$.

\begin{figure}[b]
\centering
\includegraphics[width=1.0\linewidth]{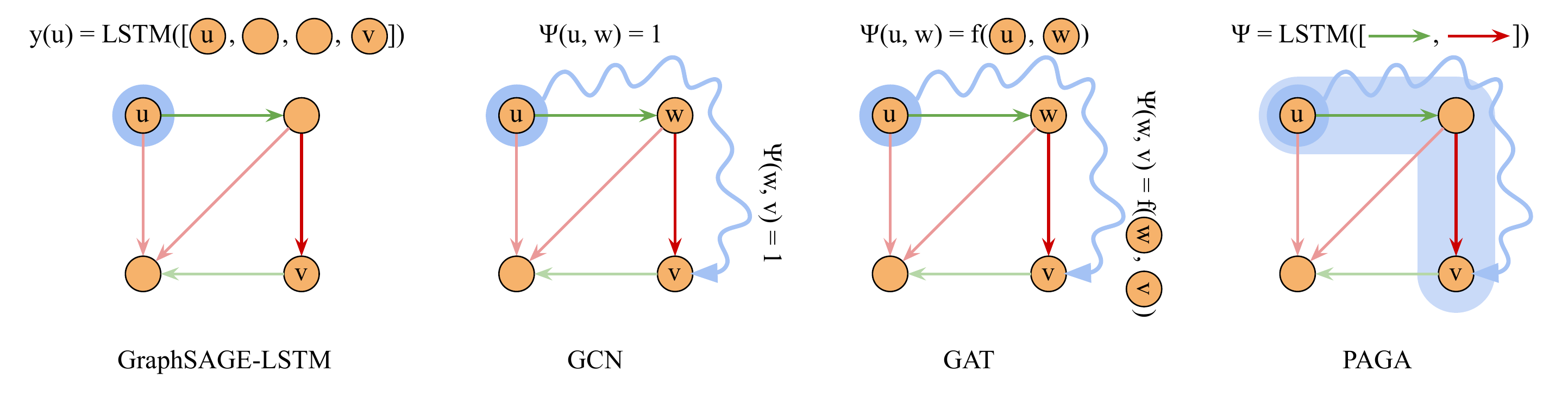}
\caption{Comparison of how $\vc{y}(u)$ is obtained from the neighborhood of $u$
in GraphSAGE (the LSTM variant), GCN, GAT and PAGA.
GraphSAGE uses LSTM to aggregate the (sampled) vertex neighborhood, while the
other three can be interpreted as attention with different $\Psi$ computation.}
\label{fig:comp_to_gcn_gat_gsage}
\end{figure}

We emphasize that this formulation of attention enables modes of vertex
interaction that cannot be easily achieved by the traditional GCN or GAT
approaches.
This will be further elaborated on in Section \ref{sec:paga_intuition} and
\ref{sec:toy_prob}. Fig. \ref{fig:comp_to_gcn_gat_gsage} compares the
structure of a few related frameworks to PAGA.

\subsubsection{Choices for Edge Sequence Feature Extractor $\Phi$}

The capability of the path-aware graph attention mechanism hinges upon the
choice of $\Phi_l$, the function that processes the type features of individual
edges along a path to determine an overall gating coefficient that filters
the influence of $v$ on $u$ via this path. Any pooling operator (such as
max-pool, avg-pool) could serve to aggregate features along the path, but the
permutation-invariance of pooling operators means they would not capture the
semantics of the ordering of edges along a path. This would be undesirable
e.g. at lane
forks, where the neighbor of the subsequent lane is not necessarily identical as
the subsequent lane of the neighbor. Therefore, a permutation-sensitive
operator, such as a recurrent neural network (in practice we use LSTM) is
potentially better. When $\lambda$ is small, a fully
connected layer (whose input width is proportional to $l$) could work as well.

\subsubsection{Efficient Implementation of Path-based Feature Extraction}

The number of possible paths from a given vertex $u$ up to length $l$ does grow
exponentially with $l$, but in practical applications on HD maps where the graph
is highly sparse, the resulting support (the $\lambda$-ring neighborhood) is
still a very small subset of $\gV$, therefore taking advantage of the sparsity
is a necessity for computation efficiency. Sparsity can be realized by storing
the vertex adjacency matrix $A$ in the \emph{compressed column/row storage}
formats, i.e. as a list of $\{u, v\}$ tuples, and structuring the graph
operations around this format.

\subsection{The Intuition Behind PAGA}
\label{sec:paga_intuition}

In essence, PAGA learns attention by trying to capture the ways that information
could flow through the graph, because it is via this flow that vertices interact
with and influence one another. Consider the middle example in Fig.
\ref{fig:road_configs} again, and let's call the two merging lanes A and B
respectively, with A being the red vehicle's lane. If the red vehicle intends to
lane change to the left, it would have to negotiate with the traffic in its
``neighbor's neighbor'' lane, i.e. the lane to the left of B (let's call
this lane C). Therefore the relationship between A and its immediate
neighbor B and second-order neighbor C is unusual: in this case the second-order
neighbor behaves like a regular neighbor lane where one would check for lane
change safety, while the immediate neighbor behaves more like the current lane
where one would look for leading vehicles and obstacles. This poses a challenge
for
the map encoder model which is responsible for discovering these high order
connections from the graph structure. Each path connecting two vertices
corresponds to a way the two lanes could come into interaction, and as witnessed
by this shallow merge example, if vertex A interacts with vertex B which in turn
interacts with vertex C, the interaction between A and C cannot be treated
recursively, as it is not necessarily an attenuated version of the interaction
between A and B or that between B and C: for example, for evaluating lane change
safety from lane A, we should put 100\% attention on C, and 0\% attention on B.
This is the intuition that motivates us to develop a path-aware mechanism for
computing attention. As investigated in Section \ref{sec:toy_prob}, a simple
toy problem with a graph exactly like this turns out to be difficult to existing
techniques like GCN, but not to PAGA.

\section{Experiments}

\subsection{Didactic Problem: Learning a Skip Interaction}
\label{sec:toy_prob}

In the same spirit as the successful practice of decomposing large convolutional
kernels into multiple layers of small (3x3) ones in computer
vision, most graph convolution methods do not consider vertices beyond the
immediate 1-ring neighborhood, and rely on stacking layers to push the
receptive field to cover long range interactions. However, we argue that such
decomposition may limit the network's expressive power for long range
interactions. As discussed above, modeling complex high order connectivity in
certain situations requires the ability to encode an interaction in a
non-recursive way. In this section we describe a simple problem constructed
to illustrate this phenomenon.

\begin{figure}[ht]
%\myvspace{-7mm}
\centering
\begin{minipage}[b]{0.18\linewidth}
\centering
\includegraphics[width=0.99\linewidth]{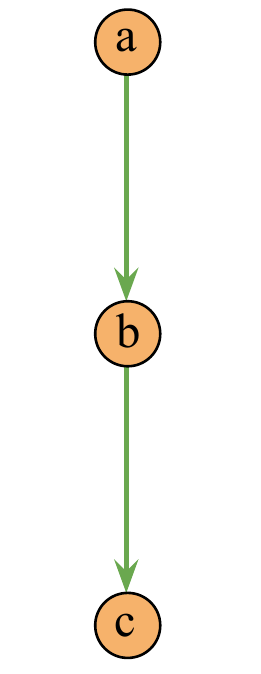}
\end{minipage}
\begin{minipage}[b]{0.8\linewidth}
\centering
\includegraphics[width=1.0\linewidth]{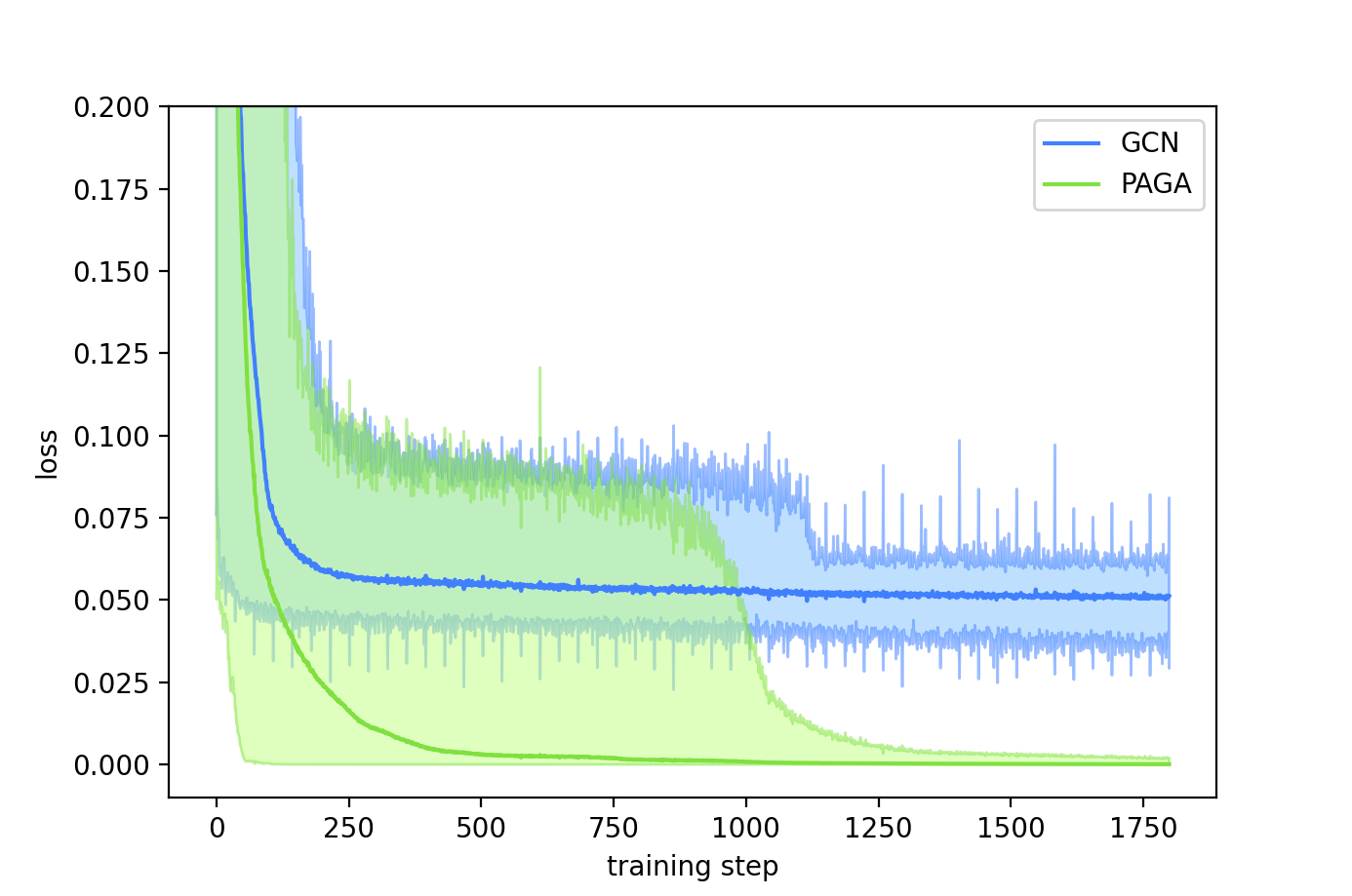}
\end{minipage}
\captionsetup{justification=centering}
\caption{Left: Graph used in the didactic problem \emph{Skip Interaction}. Right:
Loss convergence, over 100 trials.}
\label{fig:toy_prob}
%\myvspace{-3mm}
\end{figure}

Consider a graph consisting of three vertices $\gV = \{a, b, c\}$, and two edges
$\gE = \{(a, b), (b, c)\}$ (Fig. \ref{fig:toy_prob} Left). Define a function
$x: \gV \to \mathbb{R}$ as the ``feature'' on the vertices, and another function
$y: \gV \to \mathbb{R}$ as the ``label''. The task is simply learning the
functional that maps $x$ to $y$, given a prescribed construction of $y$. To
reflect the situation in the middle example of Fig. \ref{fig:road_configs}, we
define $y$ as $y(a) = x(c), y(b) = x(b), y(c) = x(c)$.
This is equivalent to the following attention $\Psi$ per (\ref{eq:att}):
\begin{align}
\Psi = \begin{pmatrix} 0 & 0 & 1 \\ 0 & 1 & 0 \\ 0 & 0 & 1 \end{pmatrix}.
\end{align}
In other words, we would like to learn a model that focuses its attention
completely on vertex $c$ for vertex $a$, while for vertex $b$ and $c$ simply
focuses on themselves. Note that the attention from $a$ to $c$ has to go through
the path $a-b-c$ topologically, but $y(b)$ must not be affected.

We generate a dataset of 4500 examples with random $x$ values on $b$ and $c$ (we
set $x(a)$ to zero, which has no effect), and train a simple implementation of
GCN and PAGA, both with hidden state size 1 and with no nonlinearities.
Removing the nonlinearity helps reduce the randomness across runs, and the
result should still be meaningful. The networks are trained on the MSE loss on
$y$ for 50 epochs with the Adam optimizer at 0.01 learning rate, and the
resulting models are evaluated on another 500 examples.

Fig. \ref{fig:toy_prob} Right shows the convergence results of the two models
on this problem over 100 trials. GCN invariably fails to converge to zero loss,
with the final loss being 0.05 on average, and MSE on $a$ near 0.1. PAGA's
training loss and evaluation MSE converge to below $10^{-3}$.

\subsection{Experimental Evaluation on Argoverse}

We then evaluate PAGA on a large scale motion prediction benchmark, the
Argoverse Motion Forecasting dataset \cite{argoverse}. The dataset comes with
detailed HD maps in the vectorized format, covering two urban areas in
Pittsburgh and Miami. The Argoverse dataset consists of over 200k
five-second-long training scenarios labeled at 10Hz, and the task is predicting
the motion of one specially marked object, called the \emph{agent}, over the
last 3 seconds of the scenario, given its motion and environment in the first 2
seconds. Prediction results are evaluated on standard trajectory distance
metrics such as ADE (average displacement error, i.e. the average of the
Euclidean distances between the predicted trajectory and the ground truth points)
and FDE (final displacement error, i.e. the Euclidean distance at the last
point), and multi-modal predictions are encouraged by the minADE$_K$ and
minFDE$_K$ metrics (ADE and FDE of the best one out of $K$ guesses output by the
predictor). To promote the predictor's awareness of the uncertainty in
its output modes, Argoverse recently introduced the brier-minADE$_K$ and
brier-minFDE$_K$ metrics, which add a penalty term on top of minADE$_K$ and
minFDE$_K$ based on the probability of the best guess output by the predictor.

In a motion prediction model, our proposed Path-Aware Graph Attention network
works as a module that parses the HD maps to provide guidance to the agent
trajectory decoder. We plug our PAGA module into the framework of LaneGCN \cite{lanegcn}
as a drop-in replacement of its map processing network based on graph
convolution. The other components of LaneGCN, such as the four-way attention
between the agents and the map, and the trajectory encoder/decoder, do not need
any modification to work with the PAGA map encoding net.

Due to the different structure of PAGA from graph convolution, some hyper
parameters we use in the experiments on Argoverse are not found in LaneGCN or
are modified, as detailed below. The attention gates are implemented as a
standard multi-headed attention with $8$ heads. We use a $\lambda$ of 2, based
on the intuition that most of the motivating examples (Fig. \ref{fig:road_configs})
involve path-dependent interactions in the 2-ring. To
reduce the branching factor, we only retain 3 out of the original 6 scales when
connecting sequential lane vertices; the lost receptive field can be easily
recovered by the built-in nonlocal interaction of PAGA. We use $C_x = C_y = 64$
channels in the vertex features, $C_e = 32$ channels in the edge features, and
we augment the edge features with the raw features (position and direction) of
incident vertices. We use $64$ channels in the hidden states of the LSTM module.
Training is run on Nvidia V100 GPUs (16GB), on servers with
Intel Xeon(R) Platinum 8163 CPUs with 336GB RAM.

Table \ref{tab:argoverse_val_vectorized} summarizes the performance of our model
compared with some state of the art methods, with the lower half of the table
focusing on vectorized representations. Our entry into the 2021 Argoverse Motion
Forecasting Competition ranked in the first place by the official metric
brier-FDE$_6$, and surpassed the other vectorized approaches by a large margin
(Table \ref{tab:argoverse_test}).

\begin{table}
\caption{Comparison to state of the art prediction techniques on the Argoverse validation set.}
\centering
\label{tab:argoverse_val_vectorized}
\begin{tabular}{lrrrrr}
\toprule
Model                                    & FDE$_6$ & ADE$_6$ & MR$_6$@2m & FDE$_1$ & ADE$_1$ \\
\midrule
MultiPath \cite{multipath}\footnote{As reported by \cite{tnt}.} & 1.68   & 0.80  & 14\%  & -    & - \\
\cite{datf}                              & 1.12   & 0.73  & -     & -    & - \\
TPCN \cite{tpcn}                         & 1.15   & 0.73  & 11\%  & 2.95 & 1.34 \\
\midrule
LaneGCN \cite{lanegcn}                   & 1.09   & 0.71  & 11\%  & 2.97 & 1.35 \\
LaneRCNN \cite{lanercnn}                 & 1.19   & 0.77  & \bf{8\%} & \bf{2.85} & 1.33 \\
VectorNet \cite{vectornet}               & -      & -     & -     & 3.67 & 1.66 \\
TNT \cite{tnt}                           & 1.29   & 0.73  & 9\%   & -    & -    \\
\midrule
\bf{Ours}                                & \bf{1.02}   & \bf{0.69}   &        & 2.87   & \bf{1.31} \\
\bottomrule
\end{tabular}
%\myvspace{-3mm}
\end{table}

\begin{table}
\caption{Experimental results on the Argoverse test set (2021 Competition final standing).}
\centering
\label{tab:argoverse_test}
\begin{tabular}{lrrrrr}
\toprule
Model                                    & brier-FDE$_6$ & FDE$_6$ & ADE$_6$ \\
\midrule
Baseline (NN) \cite{argoverse}           & 3.98 & 3.29 & 1.71 \\
\midrule
LaneGCN \cite{lanegcn}                   & 2.06 & 1.36 & 0.87 \\
LaneRCNN \cite{lanercnn}                 & 2.15 & 1.45 & 0.90 \\
VectorNet \cite{vectornet}               & -    & -    & -    \\
TNT \cite{tnt}                           & -    & 1.54 & 0.94 \\
\midrule
DM (2nd)                                 & 1.77 & \bf{1.14} & 0.81 \\
poly (3rd)                               & 1.79 & 1.21 & \bf{0.79} \\
HIKVISION (4th)                          & 1.82 & 1.19 & 0.82      \\
\midrule
\bf{Ours (1st)}                          & \bf{1.76} & 1.21 & 0.80 \\
\bottomrule
\end{tabular}
%\myvspace{-3mm}
\end{table}

\subsection{Ablation Experiments}

We report ablation experiments that are evaluated on the Argoverse validation
set. For these experiments, we train and evaluate on a decimated dataset
created by retaining one example for every ten. This drastically reduces the
training time allowing us to run multiple experiments for each setup and report
error bars; we empirically find
that the performance on the decimated dataset correlates well with that on the
full Argoverse dataset.

\paragraph{Component ablation} First of all, we would like to evaluate the
contribution of the map encoding components in the whole prediction pipeline,
which provides performance lower bounds for the other ablation studies.
We experiment with removing either the entire the Map net, or the attention
mechanism alone within
the Map net, and summarize the results in Table \ref{tab:abl_com}.
Removing the ``Map net'' component is done by deleting the entire PAGA-based map
encoder network, whose output is fused with agent encoder outputs in the LaneGCN
framework. Removing ``Attention'' is done by manually overwriting all attention
gates to zero after they are computed. It can be seen that attention provides
a small but statistically significant improvement in performance.

\begin{table}
%\myvspace{-4mm}
\caption{Component ablation study.}
\centering
\label{tab:abl_com}
\begin{tabular}{ccrr}
\toprule
\multicolumn{2}{c}{Components} \\
\cmidrule(r){1-2}
Map net & Attention & ADE$_6$ & FDE$_6$ \\
\midrule
           &            & 0.8234 $\pm$ 0.0047 & 1.3443 $\pm$ 0.0144 \\
\checkmark &            & 0.7697 $\pm$ 0.0012 & 1.1962 $\pm$ 0.0037 \\
\checkmark & \checkmark & 0.7611 $\pm$ 0.0011 & 1.1833 $\pm$ 0.0038 \\
\bottomrule
\end{tabular}
%\myvspace{-3mm}
\end{table}

\paragraph{Path attention features} The attention feature extractor function
$\Phi$ takes as input a sequence of edge
features, which include both the edge type and the spatial features (position
and direction) of the incident vertices of the edge. Table \ref{tab:abl_eft}
shows that eliminating both together degrades the prediction performance a lot
(equivalent to eliminating all attention), but keeping either one
alone shows a much smaller performance drop, suggesting that they may contain
redundant information. This can be understood intuitively, as lane connectivity
and lane spatial relationship both describe the same semantic relationship as
perceived by the drivers.

\begin{table}
\caption{Ablation of edge features in $\Phi$}
\centering
\label{tab:abl_eft}
\begin{tabular}{ccrrrr}
\toprule
\multicolumn{2}{c}{Edge Features} \\
\cmidrule(r){1-2}
Edge type & Spatial features & ADE$_6$ & FDE$_6$ \\
\midrule
           &            & 0.7711 $\pm$ 0.0015 & 1.2106 $\pm$ 0.0029 \\
\checkmark &            & 0.7617 $\pm$ 0.0001 & 1.1942 $\pm$ 0.0028 \\
           & \checkmark & 0.7621 $\pm$ 0.0021 & 1.1746 $\pm$ 0.0035 \\
\checkmark & \checkmark & 0.7611 $\pm$ 0.0011 & 1.1833 $\pm$ 0.0038 \\
\bottomrule
\end{tabular}
%\myvspace{-3mm}
\end{table}

\paragraph{Attention feature extractor $\Phi$} For the feature extractor $\Phi$,
permutation-sensitive functions such as LSTM or concatenation perform better
than symmetric functions such as summation, as in Table \ref{tab:abl_phi}.

\begin{table}
%\myvspace{-3mm}
\caption{Effects of the choice of function $\Phi$.}
\centering
\label{tab:abl_phi}
\begin{tabular}{crrrr}
\toprule
Form of $\Phi$ & ADE$_6$ & FDE$_6$ \\
\midrule
LSTM & 0.7611 $\pm$ 0.0011 & 1.1833 $\pm$ 0.0038 \\
summation & 0.7652 $\pm$ 0.0030 & 1.1924 $\pm$ 0.0075 \\
concatenation & 0.7639 $\pm$ 0.0015 & 1.1934 $\pm$ 0.0018 \\
\bottomrule
\end{tabular}
%\myvspace{-3mm}
\end{table}

\paragraph{Attention feature capacity} Eliminating the multi-headed attention
(using a single attention head, $N_{head}=1$) degrades performance as expected.
Since the number of channels in edge features, $C_e$, puts a limit on the
capacity to encode interaction types needed to capture complicated combinatoric
patterns, constraining it ($C_e=1$) has an even larger impact on prediction
accuracy. See Table \ref{tab:abl_afc}.

\begin{table}
\caption{Ablation on attention feature capacity.}
\centering
\label{tab:abl_afc}
\begin{tabular}{ccrrrr}
\toprule
\multicolumn{2}{c}{Hyper parameters} \\
\cmidrule(r){1-2}
$C_e$ & $N_{head}$ & ADE$_6$ & FDE$_6$ \\
\midrule
32 & 1 & 0.7649 $\pm$ 0.0019 & 1.1954 $\pm$ 0.0040 \\
1  & 8 & 0.7694 $\pm$ 0.0008 & 1.2096 $\pm$ 0.0033 \\
32 & 8 & 0.7611 $\pm$ 0.0011 & 1.1833 $\pm$ 0.0038 \\
\bottomrule
\end{tabular}
%\myvspace{-3mm}
\end{table}

\addtolength{\textheight}{-7cm}   % This command serves to balance the column lengths
                                  % on the last page of the document manually. It shortens
                                  % the textheight of the last page by a suitable amount.
                                  % This command does not take effect until the next page
                                  % so it should come on the page before the last. Make
                                  % sure that you do not shorten the textheight too much.

\section{Conclusion}

PAGA is developed with motivation from complex real world road configurations
and the traffic interaction therein, and we evaluated its effectiveness in
understanding nonlocal interactions on heterogeneous graphs, with both a
didactic problem that highlights the structure of the attention, and a large
scale motion prediction dataset with HD maps. We hope to continue exploring its
applications, especially problems involving modeling interaction through paths
with rich semantic context.

The computation efficiency of PAGA requires further investigation. As path
lengths increase, the number of paths grows exponentially with
the graph's branching factor. When the $\Phi$ function is permutation-invariant,
paths with the same source and target vertices could potentially be pooled to
reduce computation cost, but the number of vertices in the larger neighborhood
itself may grow exponentially anyway. We will be looking for ways to tackle this
complexity in future work.

%%%%%%%%%%%%%%%%%%%%%%%%%%%%%%%%%%%%%%%%%%%%%%%%%%%%%%%%%%%%%%%%%%%%%%%%%%%%%%%%

%%%%%%%%%%%%%%%%%%%%%%%%%%%%%%%%%%%%%%%%%%%%%%%%%%%%%%%%%%%%%%%%%%%%%%%%%%%%%%%%

%%%%%%%%%%%%%%%%%%%%%%%%%%%%%%%%%%%%%%%%%%%%%%%%%%%%%%%%%%%%%%%%%%%%%%%%%%%%%%%%
\section*{ACKNOWLEDGMENT}

We would like to thank Weixin Li, Runlin He and Chenxu Luo for their comments
and perspectives.

%%%%%%%%%%%%%%%%%%%%%%%%%%%%%%%%%%%%%%%%%%%%%%%%%%%%%%%%%%%%%%%%%%%%%%%%%%%%%%%%

\bibliographystyle{ieeetr}
\bibliography{paga.bib}

\begin{thebibliography}{10}

\bibitem{fan2019graph}
W.~Fan, Y.~Ma, Q.~Li, Y.~He, E.~Zhao, J.~Tang, and D.~Yin, ``Graph neural
  networks for social recommendation,'' in {\em The World Wide Web Conference},
  pp.~417--426, 2019.

\bibitem{fout2017protein}
A.~M. Fout, {\em Protein interface prediction using graph convolutional
  networks}.
\newblock PhD thesis, Colorado State University, 2017.

\bibitem{multipath}
Y.~Chai, B.~Sapp, M.~Bansal, and D.~Anguelov, ``Multipath: Multiple
  probabilistic anchor trajectory hypotheses for behavior prediction,'' {\em
  arXiv preprint arXiv:1910.05449}, 2019.

\bibitem{chauffeurnet}
M.~Bansal, A.~Krizhevsky, and A.~Ogale, ``Chauffeurnet: Learning to drive by
  imitating the best and synthesizing the worst,'' {\em arXiv preprint
  arXiv:1812.03079}, 2018.

\bibitem{multimodal_conv}
H.~Cui, V.~Radosavljevic, F.-C. Chou, T.-H. Lin, T.~Nguyen, T.-K. Huang,
  J.~Schneider, and N.~Djuric, ``Multimodal trajectory predictions for
  autonomous driving using deep convolutional networks,'' in {\em 2019
  International Conference on Robotics and Automation (ICRA)}, pp.~2090--2096,
  IEEE, 2019.

\bibitem{covernet}
T.~Phan-Minh, E.~C. Grigore, F.~A. Boulton, O.~Beijbom, and E.~M. Wolff,
  ``Covernet: Multimodal behavior prediction using trajectory sets,'' in {\em
  Proceedings of the IEEE/CVF Conference on Computer Vision and Pattern
  Recognition}, pp.~14074--14083, 2020.

\bibitem{mfp}
Y.~C. Tang and R.~Salakhutdinov, ``Multiple futures prediction,'' {\em arXiv
  preprint arXiv:1911.00997}, 2019.

\bibitem{matf}
T.~Zhao, Y.~Xu, M.~Monfort, W.~Choi, C.~Baker, Y.~Zhao, Y.~Wang, and Y.~N. Wu,
  ``Multi-agent tensor fusion for contextual trajectory prediction,'' in {\em
  Proceedings of the IEEE/CVF Conference on Computer Vision and Pattern
  Recognition}, pp.~12126--12134, 2019.

\bibitem{vectornet}
J.~Gao, C.~Sun, H.~Zhao, Y.~Shen, D.~Anguelov, C.~Li, and C.~Schmid,
  ``Vectornet: Encoding hd maps and agent dynamics from vectorized
  representation,'' in {\em Proceedings of the IEEE/CVF Conference on Computer
  Vision and Pattern Recognition}, pp.~11525--11533, 2020.

\bibitem{lanegcn}
M.~Liang, B.~Yang, R.~Hu, Y.~Chen, R.~Liao, S.~Feng, and R.~Urtasun, ``Learning
  lane graph representations for motion forecasting,'' in {\em European
  Conference on Computer Vision}, pp.~541--556, Springer, 2020.

\bibitem{lanercnn}
W.~Zeng, M.~Liang, R.~Liao, and R.~Urtasun, ``Lanercnn: Distributed
  representations for graph-centric motion forecasting,'' in {\em Proceedings
  of the IEEE/CVF Conference on Computer Vision and Pattern Recognition}, 2021.

\bibitem{original_gnn}
F.~Scarselli, M.~Gori, A.~C. Tsoi, M.~Hagenbuchner, and G.~Monfardini, ``The
  graph neural network model,'' {\em IEEE transactions on neural networks},
  vol.~20, no.~1, pp.~61--80, 2008.

\bibitem{ggnn}
Y.~Li, D.~Tarlow, M.~Brockschmidt, and R.~Zemel, ``Gated graph sequence neural
  networks,'' {\em arXiv preprint arXiv:1511.05493}, 2015.

\bibitem{patchy_san}
M.~Niepert, M.~Ahmed, and K.~Kutzkov, ``Learning convolutional neural networks
  for graphs,'' in {\em International conference on machine learning},
  pp.~2014--2023, PMLR, 2016.

\bibitem{graph_sage}
W.~L. Hamilton, R.~Ying, and J.~Leskovec, ``Inductive representation learning
  on large graphs,'' {\em arXiv preprint arXiv:1706.02216}, 2017.

\bibitem{spectral_gnn}
J.~Bruna, W.~Zaremba, A.~Szlam, and Y.~LeCun, ``Spectral networks and locally
  connected networks on graphs,'' {\em arXiv preprint arXiv:1312.6203}, 2013.

\bibitem{gcn}
T.~N. Kipf and M.~Welling, ``Semi-supervised classification with graph
  convolutional networks,'' {\em arXiv preprint arXiv:1609.02907}, 2016.

\bibitem{mpnn}
J.~Gilmer, S.~S. Schoenholz, P.~F. Riley, O.~Vinyals, and G.~E. Dahl, ``Neural
  message passing for quantum chemistry,'' in {\em International Conference on
  Machine Learning}, pp.~1263--1272, PMLR, 2017.

\bibitem{nn4g}
A.~Micheli, ``Neural network for graphs: A contextual constructive approach,''
  {\em IEEE Transactions on Neural Networks}, vol.~20, no.~3, pp.~498--511,
  2009.

\bibitem{ying2018graph}
R.~Ying, R.~He, K.~Chen, P.~Eksombatchai, W.~L. Hamilton, and J.~Leskovec,
  ``Graph convolutional neural networks for web-scale recommender systems,'' in
  {\em Proceedings of the 24th ACM SIGKDD International Conference on Knowledge
  Discovery \& Data Mining}, pp.~974--983, 2018.

\bibitem{kipf2016variational}
T.~N. Kipf and M.~Welling, ``Variational graph auto-encoders,'' {\em arXiv
  preprint arXiv:1611.07308}, 2016.

\bibitem{henrion2017neural}
I.~Henrion, J.~Brehmer, J.~Bruna, K.~Cho, K.~Cranmer, G.~Louppe, and
  G.~Rochette, ``Neural message passing for jet physics,'' 2017.

\bibitem{transformer}
A.~Vaswani, N.~Shazeer, N.~Parmar, J.~Uszkoreit, L.~Jones, A.~N. Gomez,
  L.~Kaiser, and I.~Polosukhin, ``Attention is all you need,'' {\em arXiv
  preprint arXiv:1706.03762}, 2017.

\bibitem{gat}
P.~Veli{\v{c}}kovi{\'c}, G.~Cucurull, A.~Casanova, A.~Romero, P.~Lio, and
  Y.~Bengio, ``Graph attention networks,'' {\em arXiv preprint
  arXiv:1710.10903}, 2017.

\bibitem{gaan}
J.~Zhang, X.~Shi, J.~Xie, H.~Ma, I.~King, and D.-Y. Yeung, ``Gaan: Gated
  attention networks for learning on large and spatiotemporal graphs,'' {\em
  arXiv preprint arXiv:1803.07294}, 2018.

\bibitem{spagan}
Y.~Yang, X.~Wang, M.~Song, J.~Yuan, and D.~Tao, ``Spagan: Shortest path graph
  attention network,'' {\em arXiv preprint arXiv:2101.03464}, 2021.

\bibitem{argoverse}
M.-F. Chang, J.~Lambert, P.~Sangkloy, J.~Singh, S.~Bak, A.~Hartnett, D.~Wang,
  P.~Carr, S.~Lucey, D.~Ramanan, and J.~Hays, ``Argoverse: 3d tracking and
  forecasting with rich maps,'' in {\em Proceedings of the IEEE/CVF Conference
  on Computer Vision and Pattern Recognition (CVPR)}, June 2019.

\bibitem{tnt}
H.~Zhao, J.~Gao, T.~Lan, C.~Sun, B.~Sapp, B.~Varadarajan, Y.~Shen, Y.~Shen,
  Y.~Chai, C.~Schmid, {\em et~al.}, ``Tnt: Target-driven trajectory
  prediction,'' {\em arXiv preprint arXiv:2008.08294}, 2020.

\bibitem{datf}
S.~H. Park, G.~Lee, J.~Seo, M.~Bhat, M.~Kang, J.~Francis, A.~Jadhav, P.~P.
  Liang, and L.-P. Morency, ``Diverse and admissible trajectory forecasting
  through multimodal context understanding,'' in {\em European Conference on
  Computer Vision}, pp.~282--298, Springer, 2020.

\bibitem{tpcn}
M.~Ye, T.~Cao, and Q.~Chen, ``Tpcn: Temporal point cloud networks for motion
  forecasting,'' in {\em Proceedings of the IEEE/CVF Conference on Computer
  Vision and Pattern Recognition}, 2021.

\end{thebibliography}

\end{document}